\documentclass[runningheads]{llncs}

\usepackage{eccv}

\usepackage{eccvabbrv}

\usepackage{graphicx}
\usepackage{tabularx, booktabs}
\newcolumntype{Y}{>{\centering\arraybackslash}X}

\usepackage[accsupp]{axessibility}  % Improves PDF readability for those with disabilities.

\usepackage{hyperref}

\usepackage{orcidlink}

\usepackage{multicol}
\usepackage{multirow}
\usepackage{xcolor, colortbl}
\usepackage{amssymb}% http://ctan.org/pkg/amssymb
\usepackage{pifont}% http://ctan.org/pkg/pifont
\newcommand{\cmark}{\ding{51}}%
\newcommand{\xmark}{\ding{55}}%
\definecolor{mygray}{gray}{0.9}
\usepackage{enumitem}
\newcolumntype{g}{>{\columncolor{mygray}}c}

\usepackage{amsmath,amsfonts,bm}

\usepackage{soul}

\def\eqref#1{equation~\ref{#1}}
\def\1{\bm{1}}

\def\ve{{\bm{e}}}

\def\vx{{\bm{x}}}

\def\mW{{\bm{W}}}

\DeclareMathAlphabet{\mathsfit}{\encodingdefault}{\sfdefault}{m}{sl}
\SetMathAlphabet{\mathsfit}{bold}{\encodingdefault}{\sfdefault}{bx}{n}

\newcommand{\R}{\mathbb{R}} % Reals
\newcommand{\mbE}{\mathbb{E}} % Eucliden
\newcommand{\cD}{{\cal D}}

\newcommand{\cF}{{\cal F}}

\newcommand{\eqdef}{:=}

\newcommand{\diag}{\rm diag}

\newcommand{\smartparagraph}[1]{\vspace{2pt} \noindent {\bf #1}}
\newcommand{\myNum}[1]{(\emph{#1})}

\newtheorem{assumption}{Assumption}
\usepackage[colorinlistoftodos,bordercolor=orange,backgroundcolor=orange!20,linecolor=orange,textsize=scriptsize]{todonotes}

\usepackage[nodisplayskipstretch]{setspace}
\usepackage{mathtools, nccmath}
\usepackage{wrapfig}
\usepackage{float}
\usepackage{makecell}

\newcommand{\res}[1]{$#1$}
\newcommand{\resb}[1]{$\mathbf{#1}$}
\newcommand{\resu}[1]{$\underline{{#1}}$}

\begin{document}

\title{Towards Multi-modal Transformers in \\Federated Learning}


\author{Guangyu Sun\inst{1}\orcidlink{0000-0002-8523-9074} \and
Matias Mendieta\inst{1}\orcidlink{0000-0002-5497-6207} \and
Aritra Dutta\inst{2}\orcidlink{0000-0001-6994-1659} \and
Xin Li\inst{2}\orcidlink{0000-0003-1201-9131} \and
Chen Chen\inst{1}\orcidlink{0000-0003-3957-7061}
}
\authorrunning{Sun et al.}

\institute{Center for Research in Computer Vision, University of Central Florida, FL, USA\and
Department of Mathematics, University of Central Florida, FL, USA\\
\email{\{guangyu, matias.mendieta, aritra.dutta, xin.li\}@ucf.edu; chen.chen@crcv.ucf.edu}}

\maketitle
\begin{abstract}
  Multi-modal transformers mark significant progress in different domains, but privacy concerns on high-quality data hinder their further improvement. Federated learning (FL) has emerged as a promising privacy-preserving paradigm for training models without direct access to the raw data held by different clients. Despite its potential, a considerable research direction regarding the unpaired uni-modal clients and the transformer architecture in FL remains unexplored. To fill this gap, this paper explores a transfer multi-modal federated learning (MFL) scenario within the vision-language domain, where clients possess data of various modalities distributed across different datasets. We systematically evaluate the performance of existing methods when a transformer architecture is utilized and introduce a novel framework called Federated modality complementary and collaboration (FedCola) by addressing the in-modality and cross-modality gaps among clients. Through extensive experiments across various FL settings, FedCola demonstrates superior performance over previous approaches, offering new perspectives on future federated training of multi-modal transformers. Code is available at \href{https://github.com/imguangyu/FedCola}{\textcolor{magenta}{https://github.com/imguangyu/FedCola}}.
\end{abstract}

\section{Introduction}\label{sec:intro}

Multi-modal transformers have led to remarkable advancements across a spectrum of downstream tasks~\cite{clip,vlmo, kang2024segvgtransferringobjectbounding}. Nevertheless, training these models demands voluminous and high-quality data
~\cite{textbooks,textbooks2}. Although high-quality training datasets contain a wealth of information, acquiring such data is resource-intensive. Moreover, many of these datasets are often closely guarded by various entities and fall under diverse privacy regulations, resulting in \emph{data silos}~\cite{kairouz2021advances}. These silos pose remarkable challenges to further improvements in the field, as they prevent the leveraging of isolated datasets to enhance model training and performance.

In response to this challenge, \textbf{federated learning} (FL)~\cite{fedavg} proposes a novel training paradigm typically designed to circumvent the need for direct access to raw data. FL enables a decentralized approach to model training. Specifically, a central server cyclically disseminates a \textit{global model} to a selected pool of clients for \textit{local training}. After the local training,  the server receives the model updates from the clients, averages them with specific weights as the \textit{global aggregation}, and applies the aggregated updates to the global model. The process continues until convergence. FL allows collaborative training on heterogenous data spread over geographically remote clients while simultaneously preserving their data privacy and offers a promising solution to the issue of data silos by facilitating the collaborative improvement of models.

\begin{figure}[t]
    \centering
    \includegraphics[width=\linewidth]{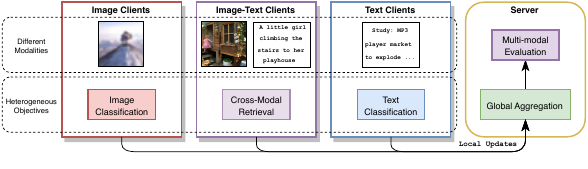}
    \caption{{The \textbf{transfer multi-modal federated learning} setting in the vision-language domain. The clients possess data of various modalities distributed across different datasets and different local training objectives. The server aims to collaboratively train a \textit{multi-modal transformer} with the data from all clients.}}
    \label{fig:setting}
\end{figure}

Current FL research predominantly addresses uni-modal scenarios and explores different aspects, such as local training~\cite{fedalign,fedprox}, personalization~\cite{tan2022towards, li2021ditto, bergou2023personalized,fedperfix, luo2023pgfed}, aggregation~\cite{fedma,fedvc}, and initialization~\cite{flpretrain,fedopt, tan2022federated}. Nonetheless, in domains with rich multi-modal data, such as healthcare and the Internet of Things (IoTs)~\cite{fediot,xiong2022unified-mmfl,pmcmfl}, \textbf{multi-modal federated learning} (MFL) are gaining attention~\cite{mmflsurvey}. 
Studies have explored horizontal MFL in healthcare, which enables training clients that share identical input modalities (\eg, all clients have text and image)~\cite{xiong2022unified-mmfl, pmcmfl,fedmsplit},
and vertical MFL in the IoTs, which enables clients that possess a single modality from the same users with multi-modal data (\eg, some clients may have text-only data, while some have their corresponding image data)~\cite{fediot}.
However, operating explicitly in either of the settings excludes other uni-modal clients with unpaired data from participating in the MFL process. Unpaired uni-modal data has shown the capability to improve the multi-modal model under a centralized setting~\cite{vlmo,flava}. 
Therefore, transferring knowledge from those uni-modal clients can further extend the scope of training data in FL, providing an important impact for MFL. But only a few studies have explored this transfer MFL setting~\cite{creamfl, liu2020federated} shown in Fig.~\ref{fig:setting}.

Although promising, there are two key challenges in the transfer MFL due to the different modality data and training objectives across clients; see Fig. \ref{fig:setting}. First, restrictive data accessibility of the 
uni-modal clients. Multi-modal clients can directly gain cross-modal knowledge from local training data, but uni-modal clients have never accessed the other modalities. Consequently, the client models become biased toward the local modality, leading to a \textit{cross-modality gap}. Meanwhile, even within the same modality, multi-modal and uni-modal models are trained with different objectives, leading to an \textit{in-modality gap}.

To tackle these challenges, we propose a novel framework for the transfer MFL in both local training and global aggregation, leveraging the unified design of transformers across different modalities. During the local training, we propose a \emph{mixture strategy} to leverage the modality-complementary information, together with a weight compression trick for efficient communication. During the global aggregation, we propose an \emph{aggregation with disaggregation} strategy to aggregate general knowledge while maintaining specific knowledge for each local objective. With these collaboration techniques for both local training and global aggregation, we build and propose a novel framework, \textbf{Fed}erated modality \textbf{Co}mplementary and col\textbf{la}boration (FedCola), providing insights for FL on multi-modal transformers, and highlight the main contributions of this paper:

\begin{enumerate}[label=(\roman*)]    
    \item \smartparagraph{Transfer multi-modal FL (\S\ref{sec:tmfl}).} To the best of our knowledge, this paper is the first to explore transformers in transfer multi-modal FL, filling the gap between centralized and federated training of multi-modal transformers.
    \item \smartparagraph{FedCola (\S\ref{sec:fedcola}).} We propose a novel framework, FedCola, to enable advanced collaboration between different modalities with only the parameters in both local training and global aggregation.
    \item \smartparagraph{Cross-modality knowledge-sharing (\S \ref{sec:local}).} We illustrate that cross-modality knowledge-sharing can be achieved without directly accessing the data but only with the model weights and provide new insights on convergence conditions of collaboration and quantified contributions.
    \item \smartparagraph{Empirical study (\S \ref{sec:experiments}).} We conduct extensive experiments on real-world datasets under different FL settings, domain gaps, and numbers of participating clients, demonstrating the effectiveness of the proposed framework.
\end{enumerate}

\section{Related Work}

\smartparagraph{Federated learning} has emerged as a powerful approach for decentralized, privacy-preserving machine learning. In practice, heterogeneity in different aspects exists among clients, including statistical heterogeneity and system heterogeneity. Most current FL focuses on addressing the statistical heterogeneity for clients with data from the same domain, the same modality, but different distributions~\cite{fedalign,fedprox, scaffold, moon, zhuang2024fedwon}. Meanwhile, the other line of work focuses on enabling edge devices with limited resources to address the system heterogeneity~\cite{feddf,fedgkt,fedgems,mendieta2024navigatingheterogeneityprivacyoneshot, Huang_2024_CVPR}. Meanwhile, FL can also be categorized by the focus of the product, which can either be the performance of the global model on more generalized test data or the performance of the client models on their local data. These two categories lead to two separate directions of FL: global and personalized FL \cite{luo2023pgfed,fedtp, fedperfix, fedbn, apfl,mortaheb2022fedgradnorm}. In this paper, we explore and address the heterogeneity in the modality level to obtain a generalized multi-modal transformer, which falls into the scope of global FL without system heterogeneity. 

\smartparagraph{Multi-modal FL.} Modality collaboration, or multi-modal fusion, allows different modalities to train and learn collaboratively by sharing high-level common knowledge and has recently been introduced in FL~\cite{feng_multimodal, xiong2022unified-mmfl, fediot}. The scope of multi-modality in FL is limited to a handful of works, and the potential is largely unexplored. For instance, earlier works consider a homogeneous model for each modality~\cite{xiong2022unified-mmfl, fediot} and further involve clients with missing modality~\cite{pmcmfl,fedmsplit}; recently, \cite{Cho2022HeterogeneousEK} took a step towards a larger server model training by knowledge transfer of diverse smaller client models, and \cite{creamfl} use knowledge-transfer between uni and multi-modal clients with heterogeneous data and learn a larger global model. We note that some multi-modal frameworks are restrictive, barring the participation of uni-modal clients \cite{xiong2022unified-mmfl}; \cite{creamfl} lifts this barrier but still relies on the auxiliary from a public dataset shared with all clients. In this paper, we focus on transferring the knowledge without any public data but with the multi-modal capability of transformers, further reducing the restrictions.

\smartparagraph{Vision transformers in FL.} Although vision transformers have become the de facto architecture in centralized computer vision tasks, their exploration under FL remains limited. Qu \etal \cite{vitfl} evaluate the performance of vision transformers under global FL settings and showcase their robustness against data heterogeneity. Meanwhile, Sun \etal~\cite{fedperfix} and Li \etal~\cite{fedtp} study the personalization of vision transformers in personalized FL. Further, research with other aspects of vision transformers (\eg., pre-trained transformers) have also gained attention~\cite{fedpeft, fedopt,flpretrain, zhuang2024foundationmodelmeetsfederated}. With a different perspective in this paper, we focus on the multi-modal capability and unified architecture of the transformer, paving the path for further large federated multi-modal transformers.
\section{Transfer MFL}\label{sec:tmfl}
We consider a transfer multi-modal FL setting in the vision-language domain with a total of $N$ clients. 
That is, we have $N_\text{v}$ image clients, $N_\text{l}$ text clients, and $N_{\text{vl}}$ image-text clients, with $N_\text{v}+N_\text{l}+N_\text{vl}=N$. Clients with the same modality own non-IID local training data with the same local objective. Uni-modal clients conduct classification tasks, while multi-modal clients conduct cross-modal retrieval tasks as their local objectives. The server owns a hold-off multi-modal test set for evaluation.

At the beginning of the FL process, the server initializes global models, $w^{(0)}={\left(w^{(0, \text{v})}, w^{(0, \text{vl}_\text{v})}, w^{(0, \text{vl}_\text{l})}, w^{(0,\text{l})}\right)}^{\top}$ for each modality combination. The entire process runs for $T$ global communication rounds. In each round, $t$, for each modality type $M$ with a sample rate $r_M$, a total of $r_MN_M$ is selected to perform the local training. Note that if all modality combinations share a fixed uniform sample rate, then $r_M=r$. Each selected client, $i$, downloads the global models, $w^{(t)}$, performs training for $E$ local epochs,
and sends the local update, $\nabla w^{(t+1)}_i = w^{(t+1)}_i - w^{(t)}$ to the server.

The server, first, aggregates the updates for each modality combination separately with a uni-modal aggregation algorithm. When FedAvg is used, the global updates for modality type $M$ is
\[
\textstyle{\nabla \bar{w}^{(t+1, M)} = \sum_{i=1}^{r_MN_M} \frac{n_i}{n_M}\nabla w^{(t+1)}_i},
\]
where $n_M=\sum_{i=1, \mathcal{M}(i)=M}^{r_MN_M}n_i$ 
is the total training samples on selected clients and $\mathcal{M}$ is a mapping that maps the client index $i$ to its corresponding modality type. 
Additionally, in our setting, we perform further collaboration across all modality types to get collaborated updates $\nabla w^{(t+1)}$ as 
\[
\textstyle{\nabla w^{(t+1)} = {\Omega}\nabla \bar{w}^{(t+1)},}
\]
where $\mathbf{\Omega}$ is the collaboration matrix for all parameters in all models $w$. 
Finally, we update the global models by $w^{(t+1)} = w^{(t)} + \nabla w^{(t+1)}$ that goes in for the next round. After $T$ rounds, we evaluate $w^{(T, \text{vl})}$ with the multi-modal test data on the server. Detailed discussion on the general transfer MFL framework and convergence guarantee in our setting are given in the supplementary material. 

\section{Method}\label{sec:fedcola}
In this section, we introduce our framework and outline two methods to mitigate cross-modality and in-modality gaps.

\begin{figure}[!t]
        \centering
        \includegraphics[width=\textwidth]{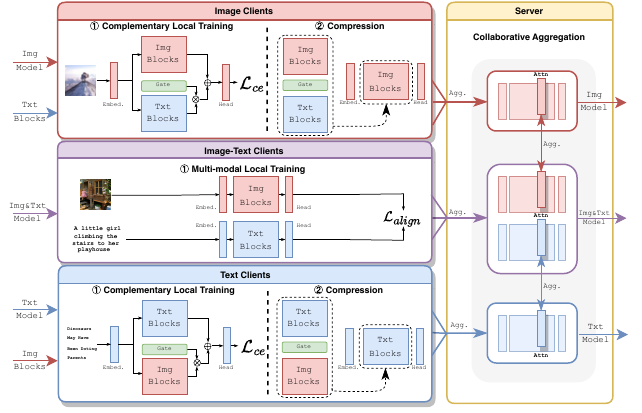}
        \caption{{Overview of our proposed framework, \textbf{FedCola}. In each round of FL, uni-modal clients download the global model for their modality along with the transformer blocks from the other modalities and perform complementary local training to address the \textit{cross-modality gap}, while multi-modal clients perform standard multi-modal local training. Then, all clients send their local updates to the server for uni-aggregation and collaborative aggregation to address the \textit{in-modality gap}.}}
        \label{fig:overview}
\end{figure}

\subsection{The framework}

To address the cross-modality and in-modality gaps, a key goal is to unify the learned representation into a shared feature space~\cite{creamfl}. Under centralized training, those gaps can be addressed by joint training in a multi-task fashion with all the data in different modalities. 
However, parameters are the only knowledge carrier in FL, due to the privacy requirements of local clients,
preventing any direct joint training across client data and tasks.
Instead, we leverage the unifying architecture of transformers across modalities and develop a novel framework with new insights into local training and global aggregation; see Fig.~\ref{fig:overview}. 
Specifically, we split the model into three parts: \myNum{i} embedding layers, \myNum{ii} transformer blocks, and \myNum{iii} task head. Among different clients, the transformer blocks are in a unified architecture, while the embedding layers and task heads are specified based on their input modalities and local training objectives. We propose a novel complementary local training pipeline on the uni-modal clients (\S \ref{sec:local}) while exploring the collaborative aggregation on the server (\S \ref{sec:agg}).

\subsection{Complementary local training against cross-modality gap}\label{sec:local}
Local training on uni-modal clients is inherently limited to utilizing only the local data available on the respective clients, thus preventing information access from other modalities. This limitation becomes particularly pronounced when employing disparate model architectures for encoding images and texts. Specifically, images cannot be encoded using the model deployed on text clients, and vice versa. Consequently, the valuable knowledge encapsulated within the text modality, which could complement the local data, remains untapped by image clients relying solely on model weights derived from other modalities.
With the adoption of transformer blocks possessing a unified architecture, tokenized local data can now be encoded using blocks originating from other modalities. This cross-modal encoding capability enables the design of paradigms to integrate signals from diverse modalities into local training processes.

As recent progress on the \textbf{Mixture of Experts} (MoE) \cite{moe, vlmo, pathway}, each layer from different expert models can be mixed by weights generated by a router, providing a new perspective for parameter collaborations. 
In contrast to the original MoE formulation, wherein all expert models are considered equally significant and the router learns to weigh them based on input characteristics, the scenario of local training on uni-modal clients presents a more straightforward approach. Given that all input data on uni-modal clients pertains to the local modality, there exists a clear emphasis on prioritizing local information, with models from other modalities serving as auxiliary components.

To signify local modality, we employ a learnable parameter initialized to 0 as the {\tt gate}, $g$. This parameter enables adaptive learning, determining how complementary knowledge from other modalities contributes to the local modality.
Formally, the output of the layer with the gate $g$ when takes $\vx$ as the input can be formulated as
$
        \mW_{\text{local}}\vx + g\mW_{\text{out}}\vx
$, where $\mW_{\text{local}}$ and $\mW_{\text{out}}$ are the local and out model weights respectively.
In this way, uni-modal clients can learn modality-complementary knowledge during local training, closing the cross-modality gap. 
To conduct complementary local training, we additionally download the transformer blocks from the other complementary modality and add gates for all linear layers within the transformer blocks, as shown in Fig.~\ref{fig:overview} (Left), encompassing self-attention layers and MLPs for the complementary cross-modal contribution.

However, if we directly send the updates of the modified local model, the upload communication cost will be almost doubled due to the involvement of more parameters. 
To remedy this, we compute the equivalent weights for each layer after the local training as $\mW_{\text{local}} + g\mW_{\text{out}}$ as a compression trick since they are linear, so the total uploading size is reduced to the size of the original model. Regarding the download size, we further explore options to reduce the required model weights from the complementary modality in \S~\ref{sec:comm_dicuss}. In this way, we provide the flexibility of the trade-off between performance and communication costs.

\subsection{Collaborative aggregation against in-modality gap}\label{sec:agg}

\begin{figure}[!t]
        \centering
        \includegraphics[width=\textwidth]{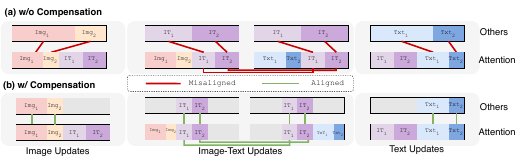}
        \caption{Illustration of the self-attention (Attention) and other layers updates with and without the proposed compensation scheme. The width of the block indicates the aggregation coefficient on that client. Without the compensation, layer-level misalignment happens between self-attention and other layers, while modality-level misalignment happens between updates of each modality on multi-modal updates. With compensation, both misalignment is fixed.}
        \label{fig:comp}
\end{figure}

Global aggregation in FL presents both opportunities and challenges. On one hand, aggregating model weights from diverse clients promises to enhance generalizability, while on the other, the inherent heterogeneity among clients poses significant obstacles to effective aggregation \cite{fedavg,xu2021deepreduce,dutta2020discrepancy}. While aggregation within the same modality constitutes a uni-aggregation problem, extensively studied within uni-modal FL, our focus shifts towards the collaboration dynamics among different modality/task combinations.

A given modality may be represented across multiple client types, \eg, image-only clients and hybrid image-text clients both use image data. The image models within these clients are trained on data from the same modality but with distinct local objectives. Consequently, the parameters of image models within these clients encapsulate the broader domain-specific knowledge inherent to image data and the intricacies of their respective local objectives.
Given that the transformer blocks adhere to a uniform architecture, direct aggregation of their model weights is feasible. However, such a uniform aggregation approach risks diluting the specific knowledge pertinent to individual local objectives. To remedy this, we propose an \textit{aggregation with disaggregation} strategy that captures overarching modality knowledge while safeguarding task-specific insights.

Inspired by transformers in centralized training paradigms~\cite{vlmo, beit}, we observe that self-attention layers encode inter-token relationships, embodying more generalized knowledge. Conversely, the subsequent MLPs, 
serve to adapt this generalized knowledge to the local objectives, thus encapsulating domain-specific knowledge.
This insight motivates us to develop the selective aggregation of updates exclusively from \textit{self-attention layers} of transformer blocks within the same modality following the aggregation of updates for each modality combination of clients. Concurrently, we advocate for the disaggregation of other components, constituting what we term \textit{in-modal collaboration}.
Formally, the in-modal collaboration matrix $\Omega$ (with compatible dimensions of the identity matrix, $I$) is given as:
\[
    \Omega^{\text{attn}}_{i,j} = 
    \begin{cases}
        \frac{n_i}{n_i+n_j}I & \mathcal{M}(i)=\mathcal{M}(j),\\
        0 & \text{otherwise},
    \end{cases}
    ;\ \Omega^{\text{others}}_{i,j} = I.
\]
Note that $\Omega^{\text{attn}}_{i,j}$ combines the self-attention layers and $\Omega^{\text{others}}_{i,j}$ combines the MLPs. 
However, such an aggregation will cause misalignment between the model weights with and without collaboration. 
Intuitively, the updates from a single client are coherent across different layers since they are directly trained. Scaling the updates for specific layers will break the coherence and lead to a misalignment in the global updates, as shown in Fig.~\ref{fig:comp}(a). 
Quantitively, during the initial uni-aggregation within the same modality type, the updates from client $i$ are equally applied to each layer of the global model with the coefficient of $\frac{n_i}{n_{\mathcal{M}(i)}}$. For the collaborative aggregation, the updates to the self-attention layers are scaled to $\frac{n_i}{n_{\mathcal{M}(i)}+n_{\text{vl}}}$, leading to \textit{layer-level misalignment}. Meanwhile, for the updates on the image-text model, self-attention layers in it are scaled with $\frac{n_\text{vl}}{n_\text{v}+n_\text{vl}}$ and $\frac{n_\text{vl}}{n_\text{l}+n_\text{vl}}$ separately, leading to a \textit{modality-level misalignment}. 

To fix such a misalignment, we propose a compensation scheme to align the updates for different layers. Specifically, we scale the updates to the non-collaborated layers with the same coefficient as the self-attention layers. With compatible dimensions of the identity matrix, $I$, and the matrix of all zeros, $0$, the compensation scheme is formulated as 
\[
\Omega_{\rm comp} = \begin{pmatrix}
        &\frac{n_\text{v}}{n_\text{v}+n_\text{vl}}I & \frac{n_\text{vl}}{n_\text{v}+n_\text{vl}}I & 0 &0 \\
        &\frac{n_{v}}{n_\text{v}+n_\text{vl}+n_\text{l}}I & \frac{n_\text{vl}}{n_\text{v}+n_\text{vl}+n_\text{l}} I & 0 &0 \\
         &0  &0  &\frac{n_\text{vl}}{n_\text{v}+n_\text{vl}+n_\text{l}}I &\frac{n_{l}}{n_\text{v}+n_\text{vl}+n_\text{l}}I\\
         &0  &0  &\frac{n_{l}}{n_\text{vl}+n_\text{l}}I &  \frac{n_{l}}{n_\text{vl}+n_\text{l}}I\\
\end{pmatrix}.
\]
Such a compensation scheme will ensure that the updates to the different parts of the transformer blocks are aligned, as shown in Fig.~\ref{fig:comp}(b). 
With in-modal collaboration, both uni-modal and multi-modal clients can gain more generalized multi-modal knowledge by aggregating from more training data. Meanwhile, the specific knowledge for their local objective is maintained by the aggregation with disaggregation strategy. Consequently, the in-modality gap is mitigated by the collaborative aggregation.

\section{Experiments}\label{sec:experiments}

 \smartparagraph{Datasets.} We follow the previous work's setting to select the datasets for each modality type~\cite{creamfl}. We use CIFAR-100~\cite{cifar} as the image data for image clients and the AG NEWS~\cite{agnews} dataset as the text data for text clients. We evaluate the performance on two different multi-modal datasets, Flickr30k~\cite{flickr30k} and COCO Captions~\cite{coco}. We sample $10,000$ images along with $50,000$ corresponding captions from Flickr30k, noted as Flickr10k, and $50,000$ image-text pairs from COCO Captions. To further evaluate the impact of the domain gap between uni-modal and multi-modal clients, we use two medical domain datasets to provide a larger domain gap. Specifically, we use OrganCMNIST~\cite{medmnist} as the medical image dataset and the Medical Abstract~\cite{medabs} dataset as the medical text dataset. Classification tasks are performed on the uni-modal clients trained with cross-entropy loss, while cross-modal retrieval tasks are performed on multi-modal clients trained with contrastive loss~\cite{contrastive}.

 \smartparagraph{FL settings.} We choose a practical FL setting as our default setting and further provide two more challenging scenarios with more heterogeneity and less participation. Specifically, we set a number of the image, text, and multi-modal clients as $12,12,8$ in our default setting. We partition the training data to corresponding clients following non-IID Dirichlet distributions with $\alpha=0.5$ without overlapping. 
In each round, $r=0.25$ of the clients on each modality type will train $5$ local epochs and 
 participate in the aggregation in a total of $T=30$ rounds. 
 For the setting with more heterogeneity, we partition the data following a Dirichlet distribution with a smaller $\alpha=0.1$, and for the setting with less participation, we use a lower participation rate of $r=0.125$.

 \smartparagraph{Model architecture.} We employ an ImageNet pre-trained ViT-Small~\cite{vit} as the transformer blocks. Images are embedded with a patch embedding layer with a patch size of 16, and texts are embedded with a BERT tokenizer~\cite{bert}. The model architecture will be the embedding layers, transformer blocks, and the classification head in the uni-modal clients, while two separate embedding layers, transformer blocks, and retrieval heads in the multi-modal clients. All the training images are resized to $224\times224$, and texts are tokenized with a maximum length of $40$.

\noindent \textbf{Comparison methods.} Considering there is no previous work focusing on the same setting, we compare our proposed method with the following methods that can apply to the transfer multi-modal FL setting:
\begin{enumerate}[label=(\roman*)]
  \item \textbf{Uni-modal methods:} We compare our proposed method with FedAvg~\cite{fedavg} and FedProx~\cite{fedprox} as the most widely applied baselines for uni-modal FL.
  \item \textbf{Adapted multi-modal methods:} Considering methods for horizontal and vertical multi-modal FL are infeasible under the transfer MFL setting, only a few multi-modal FL methods are applicable. We choose CreamFL~\cite{creamfl}, which is the state-of-the-art knowledge-distillation-based method for multi-modal FL, and FedIoT~\cite{fediot}, which extends FedAvg with a multi-modal aggregation strategy, as the comparison methods. Implementation details of the adaptation are provided in the supplementary material.
\end{enumerate}
\smartparagraph{Evaluation Metrics.} Following a related setting in previous multi-modal FL~\cite{creamfl}, we evaluate the sum of the \textit{Top-1 Recalls} on the image-to-text and text-to-image retrieval under settings with 1k and 5k test images, for which higher is better.

\setlength{\tabcolsep}{4pt}
\begin{table}[t]
  \caption{ Performance of (a) uni-modal methods, (b) multi-modal methods, and (c) our proposed method under different FL settings. Arrows mean setting changes compared to Default. The evaluation metric is the sum of top-1 \textbf{recalls} following previous settings.  
  }
  \label{tab:main}
  \centering
  \resizebox{\textwidth}{!}{
  \begin{tabular}{@{}cc>{\hspace*{4.5mm}}cc>{\hspace*{4.5mm}}cc>{\hspace*{4.5mm}}cc@{}}
    \toprule
        \multicolumn{2}{c}{\multirow{3.5}{*}{\textbf{Method}}} &\multicolumn{2}{c}{\textbf{Default}} & \multicolumn{2}{c}{\textbf{More Heterogeneity}}& \multicolumn{2}{c}{\textbf{Less Participation}}\\
    &&\multicolumn{2}{c}{ $\alpha=0.5,\ r=0.25$} & \multicolumn{2}{c}{{ $\alpha=0.1 \downarrow\ ,r=0.25$}}& \multicolumn{2}{c}{$\alpha=0.5,\ r=0.125\downarrow$}\\
    \cmidrule(lr){3-4} \cmidrule(lr){5-6} \cmidrule(lr){7-8}
    &                        &\textsc{Flickr} & \textsc{COCO}&\textsc{Flickr} & \textsc{COCO}&\textsc{Flickr} & \textsc{COCO}\\
    \midrule
    \multirow{2}{*}{\textbf{(a)}}&{FedAvg}& \res{81.08}& \res{95.42}& \res{81.70}& \res{95.32}& \res{64.82}& \res{83.91} \\
    &  {FedProx} & \res{78.55}& \res{95.16}& \res{76.33}& \res{95.62}& \res{63.33}& \res{79.88} \\
     \midrule
    \multirow{2}{*}{\textbf{(b)}}& {CreamFL} & \res{74.83}& \res{95.26}& \res{80.00}& \res{91.41}& \res{66.85}& \res{78.97} \\
       &{FedIoT} & \res{85.51}& \res{98.40}& \res{83.28}& \res{95.89}& \res{61.94}& \res{80.65} \\
        \midrule
        \textbf{(c)}&{{FedCola}} & \resb{91.96}& \resb{105.10}& \resb{91.82}& \resb{100.83}& \resb{88.85}& \resb{102.30} \\
  \bottomrule
  \\
  \toprule
        \multicolumn{2}{c}{\multirow{3.5}{*}{\textbf{Method}}} &\multicolumn{2}{c}{\textbf{More Image}} & \multicolumn{2}{c}{\textbf{More Text}}& \multicolumn{2}{c}{\textbf{Fewer Image-Text}}\\
    &&\multicolumn{2}{c}{$r=(0.33 \uparrow,\ 0.25,\ 0.25)$} & \multicolumn{2}{c}{{ $r=(0.25,\ 0.33\uparrow,\ 0.25)$}}& \multicolumn{2}{c}{$r=(0.25,\ 0.25,\ 0.125\downarrow)$}\\
    \cmidrule(lr){3-4} \cmidrule(lr){5-6} \cmidrule(lr){7-8}
    &                        &\textsc{Flickr} & \textsc{COCO}&\textsc{Flickr} & \textsc{COCO}&\textsc{Flickr} & \textsc{COCO}\\
    \midrule
    \multirow{2}{*}{\textbf{(a)}}&{FedAvg}& \res{78.28}& \res{97.28}& \res{79.62}& \res{96.69}& \res{61.12}& \res{75.10} \\
    &  {FedProx} & \res{79.25}& \res{95.39}& \res{77.59}& \res{94.96}& \res{59.19}& \res{73.08} \\
     \midrule
    \multirow{2}{*}{\textbf{(b)}}& {CreamFL} & \res{80.31}& \res{93.65}& \res{80.75}& \res{92.81}& \res{57.12}& \res{74.69} \\
       &{FedIoT} & \res{82.74}& \res{95.04}& \res{78.02}& \res{97.04}& \res{60.34}& \res{76.14} \\
        \midrule
        \textbf{(c)}&{{FedCola}} & \resb{91.24}& \resb{104.22}& \resb{90.10}& \resb{100.96}& \resb{85.68}& \resb{94.40} \\
  \bottomrule

  \end{tabular}
  }
\end{table}

\subsection{Evaluation under different FL settings}

As shown in Table~\ref{tab:main}, we first evaluate the performance of all methods under the default setting. Consistent with previous exploration on transformers in FL~\cite{vitfl}, FedAvg is a strong baseline when transformers are applied as the backbone. Meanwhile, FedProx, which has shown effectiveness with CNNs, is less effective than FedAvg with transformers. Similarly, CreamFL and FedIoT show a performance gain compared with FedAvg only under certain settings. 
However, our proposed method, FedCola, consistently surpasses all other methods across diverse settings and datasets. FedCola's incorporation of complementary modalities in local training enables sharing knowledge embedded within model weights across varied clients in FL.
Simultaneously, through collaborative aggregation, the generalizability of multi-modal models is bolstered by leveraging collaboration with uni-modal counterparts. We explore each component further in \S~\ref{sec:abl}. This approach concurrently diminishes the cross-modality gap among distinct uni-modal clients by aggregating them with multi-modal models exhibiting superior alignment, thereby yielding substantial performance improvements.

We then evaluate the performance of all methods under the settings with more heterogeneity and less participation. As shown in Table~\ref{tab:main} (Top), our proposed method, FedCola, still outperforms all other methods and does not undergo a significant performance drop. This result demonstrates the robustness of our proposed method in handling various FL settings.

\begin{figure}[!t]
    \centering
      \begin{minipage}[b]{0.5\linewidth}
      \centering
    \includegraphics[width=\linewidth]{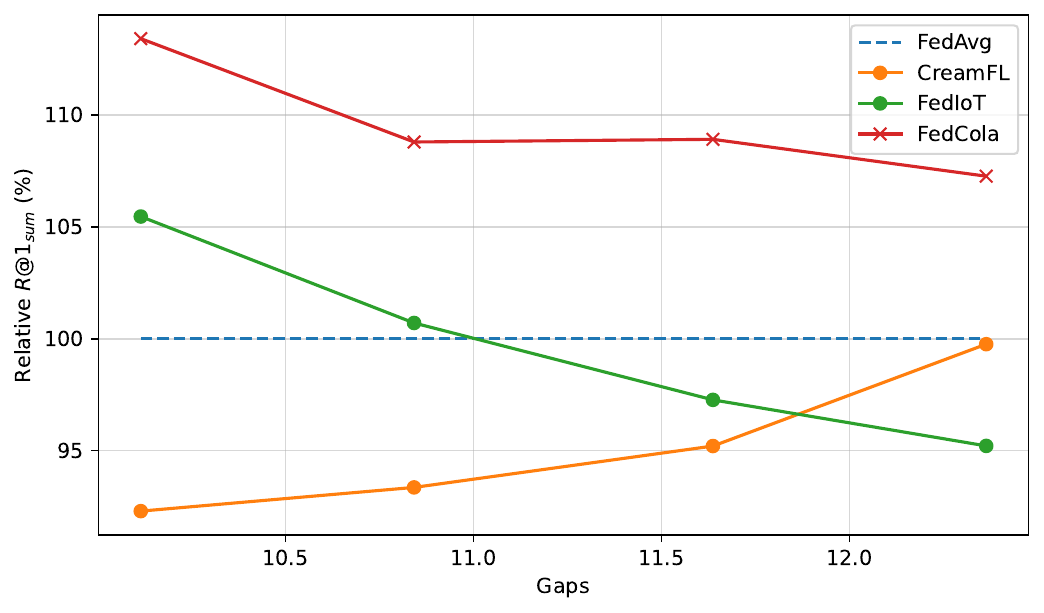}
    \caption{ Relative performance of each multi-modal method compared to FedAvg under different domain gaps}\label{fig:gap}
  \end{minipage}
    \hfill
    \begin{minipage}[b]{0.48\linewidth}
    \setlength{\tabcolsep}{4pt}
     \begin{table}[H]
      \centering
      \caption{ Impact of each proposed module. CA: collaborative aggregation; CP: compensation scheme in CA; CL: complementary local training.}\label{tab:ablation}
        \resizebox{0.7\linewidth}{!}{
        \begin{tabular}{cccc}
             \toprule
             \textbf{CA}&\textbf{CP}&\textbf{CL}& \textbf{R$@1_{\text{sum}}$}\\
             \midrule
             \xmark&\xmark&\xmark&\res{81.08} \\   
             \midrule   
             \cmark&\xmark&\xmark&\res{88.70}\\
             \cmark&\cmark&\xmark&\res{90.09}\\
             \cmark&\cmark&\cmark&\res{91.96}\\
            \bottomrule
        \end{tabular}
        }
      \end{table}
      \end{minipage}
\end{figure}

Furthermore, the number of selected clients in each modality type will affect the ratio in the aggregation. Considering the fact that uni-modal data is usually more accessible than multi-modal data, we also evaluate the performance of all methods under the setting with more participating image clients, more participating text clients, and fewer participating multi-modal clients, as shown in Table~\ref{tab:main} (Bottom). 

We notice that FedCola consistently outperforms all other methods under all settings, demonstrating the robustness of our proposed method in handling imbalanced aggregation with a consistent performance gain compared to the FedAvg baseline. 

\subsection{Evaluation under different domain gaps}

Considering different datasets are used for different modality types, it is important to evaluate the impact of domain gaps. Anchored by the multi-modal dataset, we further evaluate the robustness of the multi-modal performance when larger domain gaps are introduced. Specifically, we use the OrganCMNIST~\cite{medmnist} dataset and the Medical Abstracts~\cite{medabs} dataset to replace the image and text datasets on uni-modal clients. 

To quantify the domain gap, we use a well-used foundation model, CLIP~\cite{clip}, to extract the averaged feature embeddings for each dataset. The domain gap $g$ between the embeddings of the datasets $\ve_1$ and $\ve_2$ is then computed by the norm of the difference between the embeddings as $g=\|\ve_1-\ve_2\|_2$. The total domain gaps in a setting will be the sum of gaps in both modalities.

We choose four settings to evaluate the performance of all methods under different domain gaps: 1)~the default setting, where $g_1=10.11$, 2)~the setting with more text gap by replacing AG NEWS with Medical Abstracts, where $g_2=10.84$, 3)~the setting with more image gap by replacing CIFAR-100 with OrganCMNIST, where $g_3=11.64$, and 4)~the setting with more gaps on both modalities, where $g_4=12.36$. 

We investigate the performance of multi-modal methods compared to the FedAvg baseline relatively under these settings, and the result is shown in Fig.~\ref{fig:gap}. We notice that FedCola consistently outperforms all other methods under all settings, demonstrating the robustness of our proposed method in handling various domain gaps. However, larger domain gaps do have a negative impact on the performance of FedCola, and we leave addressing the challenge of large domain gaps as future work.
Similarly, FedIoT struggles the most in the setting with more gaps since its collaboration is based on the capability of multi-modal clients. When the domain gap is large, the multi-modal clients are not able to provide effective information to the uni-modal clients, leading to a performance drop. CreamFL, on the other hand, shows a more stable performance under different domain gaps since their collaboration is based on the public dataset, which is less affected by the domain gap despite being unable to consistently outperform the uni-modal FedAvg baseline.

\section{Discussion}
\begin{figure}[!t]
    \begin{minipage}{0.45\linewidth}
        \begin{table}[H]
      \centering
      \caption{ Communication and performance trade-off with different complementary local learning strategies. Co-Layer indicates the collaborated layers.}    
      \label{tab:cross_discuss}
      \resizebox{\linewidth}{!}{
      \begin{tabular}{cccc}
           \toprule
           \textbf{Co-Layer}&\textbf{Trainable}&\textbf{Comm. Cost (MB)} & \textbf{R$@1_{\text{sum}}$}\\
           \midrule
           None   &  - & \resb{208.81} & \res{90.09}\\
           Blocks & \cmark & \res{371.26} & \resb{91.96}\\
           Blocks & \xmark& \res{371.26} & \res{91.22}\\
           \midrule
           MLP & \cmark& \res{316.98} & \res{91.32}\\
           Attention & \cmark& \resu{262.95} & \resu{91.73}\\
          \bottomrule
      \end{tabular}
      }

  \end{table}
    \end{minipage}
  \hfill
  \begin{minipage}{0.5\linewidth}
    \begin{figure}[H]
      \includegraphics[width=\linewidth]{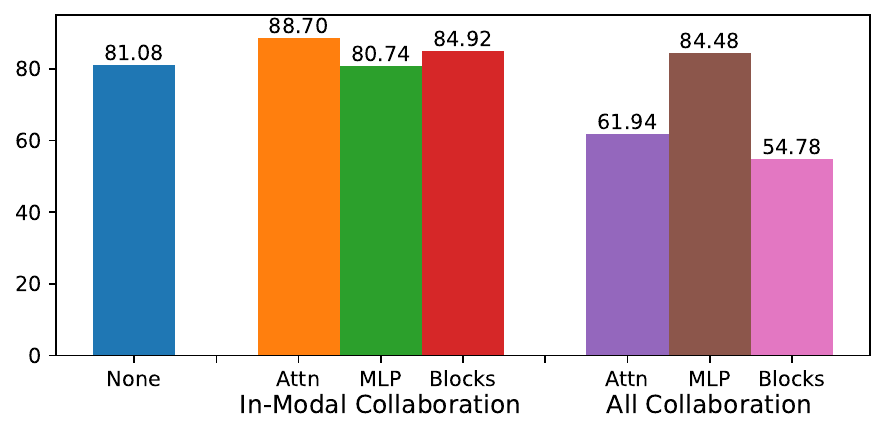}
      \caption{ Performance of different collaborative aggregation strategies}\label{fig:in_discuss}
      \end{figure}
  \end{minipage}
\end{figure}
\begin{figure}[!t]
\begin{minipage}{0.5\linewidth}
    \begin{table}[H]
        \centering
        \caption{ Performance on each test dataset.}\label{tab:uni}
        \resizebox{0.95\linewidth}{!}{
        \begin{tabular}{cccc}
             \toprule
            \textbf{Method}&\textbf{CIFAR-100}&\textbf{AG NEWS}&\textbf{Flickr}\\
            \midrule
    
             {FedAvg} & \res{89.20}& \res{87.01}& \res{81.08}\\
        
            {FedProx} & \res{88.79} & \res{85.75} & \res{78.55}\\
            \midrule
            {CreamFL} & \res{88.81} & \res{83.99} & \res{74.83}\\
            {FedIoT} & \res{87.68} & \res{83.95} & \res{85.51}\\
            \midrule
            {FedCola} & \resb{89.23} & \resb{87.07} & \resb{91.96}\\
            \bottomrule
        \end{tabular}
        }

    \end{table}   
\end{minipage}
\hfill
\begin{minipage}{0.4\linewidth}
    \begin{table}[H]
        \centering
        \caption{ Scaling-up capability with more uni-modal datasets.}\label{tab:scaling}
        \resizebox{\linewidth}{!}{
        \begin{tabular}{cccc}
             \toprule
            \textbf{\# Datasets }&0 & 1& 2\\
            \midrule
            $R@1_{\text{sum}}$ & \res{81.08} &\res{91.96} & \res{93.25}\\
            \bottomrule
        \end{tabular}
        }
    \end{table}
    
\end{minipage}
\end{figure}

\subsection{Ablation study} \label{sec:abl}

\smartparagraph{Impact of each module.} As shown in Table~\ref{tab:ablation}, we observe that when the in-modal collaboration is applied, the performance is improved to $88.70$ by the better generalizability gained from the collaborated self-attention layers. By further addressing the misalignment between layers and modalities, the performance is further improved to $90.09$. Finally, with the cross-modal collaboration on the local training, each uni-modal model learns from the other modalities, improving the performance to $91.96$.

\smartparagraph{Designs for complementary local training.}\label{sec:comm_dicuss} For the out-modality model weights, they can be either trainable or frozen during the local training. As shown in Table~\ref{tab:cross_discuss}, the trainable weights from the other modalities are more effective than the frozen out-modality weights, indicating that the collaboration is more effective when out-modality weights are updated during the local training. 
Meanwhile, considering introducing such out-modality weights will increase the downloading cost, so we propose alternative strategies, which are more communication-efficient. We notice that although leveraging the entire transformer blocks is more effective for complementary local training, performing it with the self-attention layers only provides a better trade-off between performance and communication cost.

\smartparagraph{Designs on collaborative aggregation.} During the collaborative aggregation, we choose to perform the collaboration on the self-attention layers between the same modalities (\ie, in-modal collaboration). We also study the alternative strategy for collaborative aggregation. The collaboration can be performed on MLPs or the entire transformer blocks, and the collaboration range can be extended to all models instead of the same modalities as the model architectures are the same (\ie, all collaboration). As shown in Fig.~\ref{fig:in_discuss}, the self-attention layers are the most effective for collaboration, supporting that the self-attention layers carry more general and collaborative knowledge. Meanwhile, the collaboration between the same modalities is more effective than the collaboration between all models, indicating that the cross-modality gap is more significant than the in-modality gap. We leave the exploration of the cross-modal collaboration on the global aggregation as future work.

\subsection{Fairness analysis}

Unlike centralized training, where all data is owned by one data center aiming at one objective \cite{dutta2020discrepancy}, the data in FL are from different clients for different local tasks. We further study the performance on uni-modal tasks and quantify the contributions of each type of client to the performance gain. Such an analysis provides insights into the fairness of the collaboration and an initial guideline for profit sharing among different clients~\cite{li2019fair,song2019profit}.

\smartparagraph{Performance on uni-modal tasks.} We evaluate the performance of each method on the uni-modal tasks. As shown in Table~\ref{tab:uni}, we observe that FedCola maintains the performance on the uni-modal tasks and improves the multi-modal performance, while the other methods sacrifice the uni-modal performance.

% \begin{figure}[!t]
% \begin{subfigure}[t]{0.32\textwidth}
%     \centering
%     \includegraphics[width=\linewidth]{figures/fedavg_crop.pdf}
%     \caption{FedAvg}
%     \label{fig:fedavg}
% \end{subfigure}
% \hfill
% \begin{subfigure}[t]{0.32\textwidth}
%     \centering
%     \includegraphics[width=\linewidth]{figures/fedcola_crop.pdf}
%     \caption{FedCola}
%     \label{fig:fedcola}
% \end{subfigure}
% \begin{subfigure}[t]{0.33\textwidth}
%     \centering
%     \includegraphics[width=\linewidth]{figures/lda.png}
%     \caption{Extracted features}
%     \label{fig:features}
% \end{subfigure}
% \caption{ Visualization of the parametric loss landscape with Hessian eigenvectors $\epsilon_0$ and $\epsilon_1$ and the extracted features for each resulting global multi-modal model.}
% \end{figure}

\smartparagraph{Quantifying the contributions of uni-modal clients.} Quantifying the contribution is a significant problem in federated learning regarding fairness. We leverage the Shapley value~\cite{shapley} to investigate the contribution of each type of uni-modal client. The Shapley value is a solution to the problem of fair distribution of the total gains generated by the coalition of players. It is the average marginal contribution of a player to all possible coalitions. In our case, the coalition is the combination of different modality clients. We consider all four possible coalitions of the participation of uni-modal clients: 1) No uni-modal clients, 2) only image clients, 3) only text clients, and 4) both image and text clients. For each coalition, we evaluate the performance on Flickr and assume the probability of each coalition is equal. The Shapley value of image clients and text clients is $4.74$ and $6.14$, respectively, indicating that the text clients contribute more to the performance gain in the default setting. Consequently, the profit sharing should be more favorable to the text clients.

\subsection{Scaling-up capability}
To verify the scaling-up capability, we study the trend of the performance on multi-modal evaluation when more uni-modal datasets (\ie, CIFAR-100, AG NEWS, OrganCMNIST, and Medical Abstracts) are involved. As shown in Table~\ref{tab:scaling}, the performance increases when more generalized data is leveraged in FL, supporting the scaling-up capability of the proposed framework. 

% \subsection{Visualization}
% The smoothness of the parametric loss space has been utilized as a significant indicator of the model generalizabilty~\cite{fedalign,chen2020stabilizing,zela2019understanding}. To illustrate that FedCola learns a more generalized global model, we visualize the loss space on $256$ training samples of FedAvg (Fig.~\ref{fig:fedavg}) and FedCola (Fig.~\ref{fig:fedcola}) when the weights of the model are perturbed along the direction of the top Hessian eigenvectors. The loss landscape of FedCola is significantly smoother than FedAvg, indicating that with the help of the proposed framework, a more generalized global model can be obtained. Additionally, we further conduct visualizations with Linear Discriminant Analysis (LDA) at the feature level, as shown in Fig.~\ref{fig:features}. By computing the distance between the feature centers, we find the gaps between uni-modal and multi-modal datasets are reduced under FedCola. 

% \vspace{-3mm}
\section{Conclusion}
% \vspace{-1mm}
% In this paper, we propose a novel framework, FedCola, under a transfer multi-modal federated learning setting with transformers. FedCola leverages complementary local training and collaborative aggregation with only model weights without any public data and outperforms previous methods under different scenarios. We conduct extensive experiments under different FL settings, domain gaps, and numbers of participating clients, demonstrating the effectiveness of the proposed framework and providing new insights including the convergence conditions of collaboration and contributions of each type of client. 
% We believe that our work will inspire more research on multi-modal federated learning towards large federated learned multi-modal transformers.
In this work, we present FedCola, a novel framework tailored for transfer MFL focussing on transformers. By integrating complementary local training and collaborative aggregation techniques that solely depend on model parameters and dispense the need for public data, FedCola demonstrates improved performance over previous approaches under diverse FL scenarios, including different domain gaps and numbers of participating clients. Our framework proves its effectiveness through extensive numerical experiments and offers new insights into the dynamics of collaboration, including conditions for convergence and the specific contributions of different client types. We envision that our contributions will encourage further investigation into multi-modal federated training paradigms, particularly to train large transformer architectures.

% \noindent\textbf{Acknowledgement.}
% \clearpage
\section*{Acknowledgement}
This work is partially supported by the NSF/Intel Partnership on MLWiNS under Grant No. 2003198.

% \clearpage  % TODO REVIEW/FINAL: This \clearpage needs to be removed from both review and camera-ready versions.

\bibliographystyle{splncs04}
\bibliography{main}

\newpage

\section*{Supplementary Material} 
\appendix

%\tableofcontents

The supplementary material is structured as follows:
\begin{enumerate}
    \item In Section~\ref{sec:app_theoretical} we sketch a general framework for transfer MFL (\S\ref{sec:general framework}), then we adopt it to the transfer MFL setting in the vision-language (\S\ref{sec:setup}). We give the convergence guarantee in \S\ref{sec:convergence}. Finally, in \S\ref{sec:distribution}, we argue why multi-modal FL might work with diverse data modalities by providing a simple heuristic on the distribution of the multimodal data.
    \item Section~\ref{sec:exp} elaborates on more experimental details, including \myNum{i} implementation details, \myNum{ii} visualization of the data partitioning, \myNum{iii} stochasticity discussion, \myNum{iv} breakdown performance of the results in the main paper, \myNum{v} communication analysis, \myNum{vi} experiments under imbalanced total client numbers, and \myNum{vii} visualizations.
    \item Section~\ref{sec:lim} discusses the potential negative social impact and limitation of the proposed method.
\end{enumerate}

\section{Theoretical Guarantee}\label{sec:app_theoretical}
In this section, we start with a general transfer multimodal FL framework. 
\subsection{A general framework for transfer MFL}\label{sec:general framework}
In this section, we outline the general algorithmic framework of transfer MFL. Consider the {\em empirical risk minimization}~(ERM) problem:
\begin{equation}\label{eq:opt}
\min_{w\in\R^d} \left[\cF(w) \eqdef \sum_{j=1}^{M}F_j(w)=\sum_{j=1}^{M}\left(\underbrace{\frac{1}{N_j}\sum_{i=1}^{N_j}f_{ji}(w)}_{:=F_j}\right)\right],
\end{equation}
where 
%$f_{ji}(w)\eqdef\frac{1}{D_n}\sum_{k=1}^{D}\mbE_{z_k \sim \cD_i}{l(w;z_k)}$ 
$f_{ji}(w)=\frac{1}{|D_j|}\sum_{i\in D_j} \mbE_{z \sim \cD_i}{l(w;z)}$
denotes the loss function evaluated on input datapoint at the $i^{\rm th}$ client, with $D_j$ denoting the set of all clients whose data has non-empty $j$th modality, $M$ is the total number of modality types, and $N_j$ denotes the number of clients for the $j^{\rm th}$ modality. At the beginning of the training process, the server initializes global models, $w^{(0)}.$ The entire process runs for $T$ global communication rounds.~At the beginning of each global round, $r$, each client, $i$, %from the respective modality 
receives the global model, initializes it to $w_i^0=w^{(r)}$, and updates its local model in each iteration, $t$. We compactly write it:
\begin{equation}\label{eq:client_update}
w_i^{t+1}=w_i^t-\Omega_{\rm local}\nabla w_i^t,
\end{equation}
where $\Omega_{\rm local}$ is a block diagonal matrix, where each block is of the size of the parameters of that individual modality type. Each client completes local training (\ref{eq:client_update}), for $E$ local epochs, communicates the model update to the server, and the server aggregates them at the $r^{\rm th}$ round via:
\begin{equation}\label{eq:server_aggr}
\nabla w^{(r)} = \Omega_{\rm Aggregation}(w_i^{E}-w^{(r)}).
\end{equation}
Finally, the update rule at the server for each round $r$ is given as 
\begin{equation}\label{eq:server_update}
w^{(r+1)}=w^{(r)}+\Omega_{\rm server}\nabla w^{(r)},
\end{equation}
where $\Omega_{\rm server}$ is a symmetric, $(M+1)\times (M+1)$ block band matrix. %This process is schematically explained in Figure \ref{fig:multimodal-schematic}. 

\subsection{Transfer MFL in our setting}\label{sec:setup}
We consider a transfer multi-modal FL setting in the vision-language domain with a total of $N$ clients. Based on the general framework, we describe our problem setup. Although (\ref{eq:opt}) presents a general multimodal loss function, in our case, we are working with 3 modality types, so, $M=3$. Let $D_\text{v}, D_\text{l},$ and $D_\text{vl}$ present the dataset for vision, text, and multimodal tasks, respectively. Note that, $D_\text{vl}=\{d:d=(v,l)\}$ contains vision and text pair as input, and define $D_\text{vl}(v):=P_v(D_\text{vl})$, $D_\text{vl}(l):=P_l(D_\text{vl}),$ where $P_y$ is the projection operator on the set $y.$ For training the vision models, we use $D_\text{v}\cup D_\text{vl}(v)$; for training the language models, we use $D_\text{l}\cup D_\text{vl}(l).$ At the beginning of the FL process, the server initializes global models, $w^{(0)}:={\left(w^{(0, \text{v})}, w^{(0,\text{l})}, w^{(0, \text{vl}_\text{v})}, w^{(0, \text{vl}_\text{l})}\right)}^{\top}$ %Denote $w=\begin{pmatrix} w_v\;w_l\;w_{vl_v}\;w_{vl_l}\end{pmatrix}^\top$, 
the set of parameters, and $\nabla w^r=\begin{pmatrix} \nabla w_v^r\;\nabla w_l^r\; \nabla w_{vl_v}^r\;\nabla w_{vl_l}^{r} \end{pmatrix}^\top.$ 

For local training, 
%$\Omega_{\rm local}=\diag(\eta_vI_{|w_v|}\;\eta_lI_{|w_v|}\;\eta_{vl}I_{|w_{vl_v}|}\;\eta_{vl}I_{|w_{vl_l}|}),$ 
$\Omega_{\rm local}=\diag(\eta_vI_{|D_v|}\;\eta_lI_{|D_l|}\;\eta_{vl}I_{|D_{vl}(v)|}\;\eta_{vl}I_{|D_{vl}(l)|}),$ 
where $\eta_v,\eta_l,\eta_{vl}\ge 0$ are stepsizes for respective modality type. Note that, $|x|$ denotes the dimension of an arbitrary vector, $x,$ and $I_{|x|}$ is the identity matrix in the space $\R^{|x|\times |x|}.$ Let $n_v,n_l$ and $n_{vl}$ clients be sampled uniformly at random from each $\{N_j\}_{j=1}^3$, respectively, in each training round. Hence, after $E$ local epochs, (\ref{eq:server_aggr}) is given as:
\begin{equation}
\begin{cases}
\nabla w_v^r:= \frac{1}{n_v}\sum_{i\in n_v}(w_{v_i}^E-w_v^r), \nabla w_l^r:= \frac{1}{n_l}\sum_{i\in n_l}(w_{l_i}^E-w_l^r),\nonumber\\
\begin{pmatrix}
 \nabla w_{vl_v}^r\\\nabla w_{vl_l}^{r}  
\end{pmatrix}:=\frac{1}{n_{vl}}\sum_{i\in n_{vl}}\begin{pmatrix}
 w^E_{{vl_v}_{i}}-w_{vl_v}^r\\ w^E_{{vl_l}_{i}}-w_{vl_l}^r  
\end{pmatrix}.
\end{cases}
\end{equation}
Our setting allows different structures of $\Omega_{\rm server}$ that can be adapted to consider cross-modal contribution. Specifically, we consider $\Omega_{\rm server}$ as follows: For $i$ odd, $(\Omega_{{\rm server}})_{ii}=A_v$, and for $i$ even, $(\Omega_{\rm server})_{ii}=B_l$. The superdiagonal blocks, $j>i$, are given as: \myNum{i} For $i$ odd and $j$ even, $(\Omega_{\rm server})_{ij}=0;$ for $j$ odd, $(\Omega_{\rm server})_{ij}=B_{vl};$ \myNum{ii} For $i$ even and $j$ odd, $(\Omega_{{\rm server}})_{ii}=0$, for $j$ even, $(\Omega_{\rm server})_{ij}=D_\text{vl}.$ 
Constructing $\Omega_{\rm server}$ as above allows us to leverage the participation and interaction of each modality over the {\em vanilla} multimodal FedAvg. For {\em vanilla} FedAvg \cite{fedavg,scaffold}, $\Omega_{\rm server}$ is a block diagonal matrix and $n_v=n_l=n_{vl}.$ 

\subsection{Convergence guarantee}\label{sec:convergence}
%We consider a transfer multi-modal FL setting in the vision-language domain with a total of $N$ clients.
Based on the convergence of FedAvg in \cite{fedavg, scaffold}, in this section, we will comment on the convergence of general transfer MFL. %for modality types, $M=3$. We consider to have $N_{v}$ image clients, $N_{l}$ text clients, and $N_{vl}$ image-text clients, with $N_{v}+N_{l}+N_{vl}=N$. 
For ease of notation we consider, at each round, $S_j$ be the number of clients sampled for the the $j^{\rm th}$ modality. 

\subsubsection{Assumptions.}\label{sec:method}
We require the following assumptions. 
\begin{assumption}\label{ass:minimum}
	\textbf{(Global minimum)} 
	For each $j\in[M]$, there exists $w^\star$ such that, $F_j(w^\star)=F_j^\star\leq F_j(w)$, for all $w\in\R^d.$  
\end{assumption}

\begin{assumption}\label{ass:smoothness}\textbf{($\beta$-Smoothness)} The loss function $f_{ij}: \R^d\to \R$ at each node is $\beta$-smooth, i.e. $f_{ij}(y)\leq f_{ij}(x)+\nabla f_{ij}(x)^\top(y-x)+\frac{\beta}{2}\|y-x\|^2$ for all $x,y\in\R^d$. 
\end{assumption}
\begin{remark}\label{remark:beta_smooth}
	The above assumption implies that $F_j$ is $\beta$-smooth for all $j$.
\end{remark}

\begin{assumption}\label{ass:noise}
    For each $j\in[M]$, there exist constants $G_j\ge 0, B_j\geq1$, such that for all $x\in \R^d$, the stochastic noise, $\xi_{i,t}$ follows  
	\begin{eqnarray*}
	\frac{1}{N_j}\sum_{i=1}^{N_j}\|\nabla f_{ij}(x)\|^2\le G_j^2+B_j^2\|F_j(x)\|^2.
	\end{eqnarray*}
\end{assumption}

\begin{assumption}\label{ass:bounded_grad}\textbf{(Bounded variance)}
  For each $j\in[M]$, let $g_{ji}(w):= \nabla f_{ji}(w,z_{i(k)})$ be the unbiased stochastic gradient of $f_{ji}$ with bounded variance. That is, there exists, $\sigma_j\ge 0$ such that,  $\mathbb{E}_{z_{i(k)}}\left[\|g_{ji}(w)-\nabla f_{ji}(w)\|^2\right]\le \sigma_j^2,$ for all $w,i$, where $z_{i(k)}$ is the $k^{\rm th}$ sample data at the $i^{\rm th}$ node. 
\end{assumption}

\begin{assumption}\label{ass:positive_eigenvalue}
The eigenvalues of the symmetric matrix, $\Omega_{\rm server}$ are nonnegative. 
\end{assumption}
\begin{remark}\label{remark:eigen_decomposition}
	The above assumption implies that there exists an orthogonal matrix $P$ such that, $\Omega_{\rm server}=P\Lambda P^\top$, where $\Lambda$ is a diagonal matrix of nonnegative eigenvalues of $\Omega_{\rm server}$. 
\end{remark}
\begin{remark}\label{remark:updated_iterate}
    Based on the previous remark, the update rule (\ref{eq:server_update}) can be rewritten as: 
    \begin{eqnarray*}
        P^\top w^{(r+1)}= P^\top w^{(r)}+P^\top P\Lambda P^\top \nabla w^{(r)}.
    \end{eqnarray*}
    Consider the change of variable, $\tilde w^{(r)}:=P^\top w^{(r)}$ and hence the above becomes:
    \begin{equation}\label{eq:server_update_modified}
    \tilde w^{(r+1)} = \tilde w^{(r)}+\Lambda \nabla \tilde w^{(r)}. 
    \end{equation}
\end{remark}

Finally, we are all set to give our main convergence result based on the vanilla FedAvg framework; for more details see \cite{fedavg, scaffold}. 

\begin{theorem}
   For each $j\in[M]$, let $F_j$ satisfies Assumptions \ref{ass:minimum}-\ref{ass:positive_eigenvalue}. Then $$\mathbb{E}\left[ \|\nabla F_j(\tilde w^{(T)})\|^2 \right] \le O\left(\frac{\beta\sqrt{(F_j(\tilde w^{(0)})-F_j^\star)}}{\sqrt{TES_j}}\right).$$
\end{theorem}

\begin{remark}
    From (\ref{eq:opt}), we have $\cF(\tilde w)=\sum_{j=1}^{M}F_j(\tilde w)$. Hence the boundedness of each $\mathbb{E}\left[ \|\nabla F_j(\tilde w^{(T)})\|\right]$ guarantee the boundedness of $\mathbb{E}\left[ \|\nabla \cF(\tilde w^{(T)})\|\right].$
\end{remark}

\subsection{Distribution of the data}\label{sec:distribution}
\begin{figure}[!t]
    %     \centering
    %     \includegraphics[width=\textwidth]{figures/overall.pdf}
    %     \vspace{-5mm}
    %     \caption{\small{Overview of our proposed framework, \textbf{FedCola}. In each round of FL, uni-modal clients download the global model for their modality along with the transformer blocks from the other modalities and perform complementary local training to address the \textit{cross-modality gap}, while multi-modal clients perform standard multi-modal local training. Then, all clients send their local updates to the server for uni-aggregation and collaborative aggregation to address the \textit{in-modality gap}.}}
    %     \label{fig:overview}
    % \end{minipage}
    % \vspace{-3ex}
    % \hfill
        
        \begin{minipage}{\textwidth}
        \centering
        \includegraphics[width=0.7\textwidth]{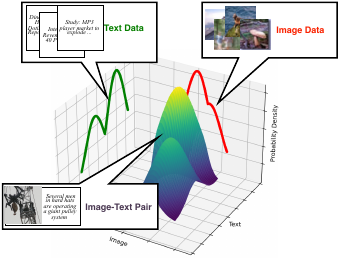}
        \caption{Multi-modal FL in the vision-language domain with collaboration from different modalities.}
        \label{fig:prob}
        \end{minipage}
        % \hfill 
        \begin{minipage}{\textwidth}
        \centering
        \includegraphics[width=0.9\linewidth]{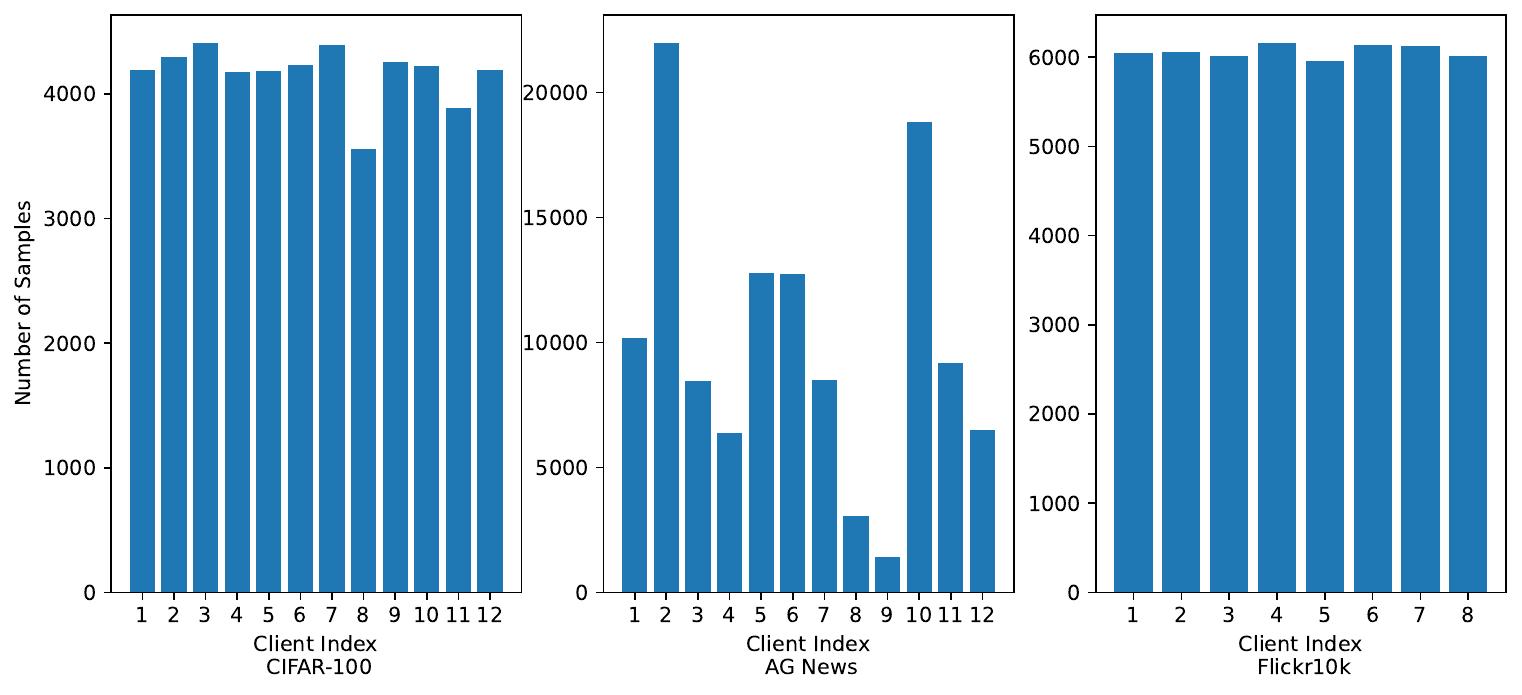}
        \caption{Visualization of the data partitioning of different datasets: CIFAR-100 \cite{cifar}, AG News \cite{agnews}, Flickr10k \cite{flickr30k}.}
        \label{fig:data}
        \end{minipage}

\end{figure}

In this subsection, we discuss the perspective of multi-modal learning from the learning of the joint distribution of the modalities. In general, we could (and should) not assume independence among the different modalities at hand, and thus, their joint distribution is not simply the product of the marginal distributions. The training datasets then should be samples that reflect the same distributions of each modality feature. Let $\cD$ be the joint {\em unknown} distribution of the input data of two modalities. Let the datasets, $D_\text{v}$ and $D_\text{l}$ have $\cD_\text{v}$ and $\cD_\text{l}$ as their marginal probability distributions of modalities $v$ and $l$ respectively. The availability of the dataset $D_\text{vl}$ that follows the distribution $\cD$ makes it possible to learn the joint distribution when $v$ and $l$ modalities are not independent; see Figure \ref{fig:prob}. The learning of a joint density model over the space of multimodal inputs is
likely to yield a better generalization in various applications \cite{potamianos2017audio}. Intuitively, this explains the possibility that multi-modal learning could improve performance when modalities are jointly used in training the parameters even for the individual modality model.    

\section{Numerical Experiments}\label{sec:exp}
This section serves as an addendum to the numerical experiments in the original paper. 

\subsection{Implementation Details}

\subsubsection{General setup.} We use AdamW~\cite{adamw} as the optimizer with a learning rate of $0.0001$ with a decay of $0.99$ every epoch for local training. The batch size is set as $112$ in most cases. All the experiments are implemented under the PyTorch framework and run on $4\times$ Nvidia A5000 GPUs. 

\subsubsection{CreamFL~\cite{creamfl}.} In the original CreamFL, there is public data in the server on which the global model can be directly trained. However, we assume no training data on the server following the traditional FL setting. Therefore, we replace the centralized training on the server with an aggregation of the client models. We use $500$ samples from the MS-COCO~\cite{coco} dataset for knowledge distillation and set the optimal distillation and local contrastive weights as $1$ and $1e-7$, respectively, after a parameter search.

\subsubsection{FedIoT~\cite{fediot}.} We follow the original design of FedIoT by applying a factor of $100$ to the multi-modal models during aggregation of the transformer blocks.

\subsection{Visualization of the data partitioning}
We perform non-IID data partitioning to simulate the client data. For CIFAR-100 and AG NEWS datasets, we partition samples of each class with a random Dirichlet distribution with a given $\alpha$. For Flickr10k, we apply a non-IID number of training samples due to a lack of class labels, following~\cite{superfed}. A visualization of the number of samples on each client is shown in Fig.~\ref{fig:data}.

\subsection{Stochasticity discussion}
To study the impact of the stochasticity in the experiments, we additionally conduct experiments with two additional random seeds besides the original seed $1$ reported in the main paper and report the standard deviation (STD) of each method. As shown in Table~\ref{tab:seed}, FedCola consistently outperforms all the comparison methods with a significant gap meanwhile holding the smallest STD, indicating the effectiveness and robustness of our proposed method.

\setlength{\tabcolsep}{5pt}

\subsection{Breakdown performance}
To provide more details for the reported results under each setting in the main paper, a breakdown performance with image-to-text top-1 recall (i2t $R@1$), text-to-image top-1 recall (t2i $R@1$) under both the 1k and 5k test image settings are given in Table~\ref{tab:flickr} for Flickr and Table~\ref{tab:coco} for COCO Captions.

\subsection{Communication analysis}
\begin{wraptable}{r}{0.5\linewidth}
    \centering
    \caption{Communication cost and performance of each method on Flickr}
    \label{tab:comm}
    \resizebox{\linewidth}{!}{
    \begin{tabular}{ccc}
        \toprule
         \bf Method& \bf Comm. Cost (MB) &\bf $\mathbf{R@1_\text{sum}}$  \\
         \midrule
         FedAvg&\res{208.81}&\res{81.08}\\
         FedProx&\res{208.81}&\res{78.55}\\
         CreamFL& \res{211.74}&\res{74.83}\\
         FedIoT&\res{208.81}&\res{85.51}\\
         \midrule
         FedCola (CA-only)&\res{208.81}& \res{90.09}\\
         FedCola (Attn)& \res{262.95} & \res{91.73}\\
         FedCola& \res{371.26}& \res{91.96}\\
         \bottomrule
    \end{tabular}
    }
\end{wraptable}
In \S 6.1 in the main paper, we study the communication trade-off of the proposed complementary local training. We further propose the communication costs of the comparison methods as a reference. Specifically, we report the size of the total download communication on one image client and one text client. An extended version is shown in Table~\ref{tab:comm}. It shows that even FedCola with collaborative aggregation only (CA-only) can outperform all comparison methods without additional communication overhead. Further, when more communication budget is acceptable, FedCola (Attn) can provide a better trade-off between communication cost and performance, while the original FedCola can provide the highest performance.

\subsection{Imbalanced client scenario}
In the main paper, we reported the performance when the number of participating clients is imbalanced and the number of total clients is the same as the default setting, considering the total client numbers in each type of client will only impact the uni-aggregation before the collaboration. To provide more experimental results, we report the performance under there are more image clients ($N_\text{v}=16$ increased from $12$) and more text clients ($N_\text{l}=16$ increased from $12$) in Table~\ref{tab:imb}. As expected, FedCola still outperforms all comparison methods under such settings.

\begin{figure}[!t]
\begin{minipage}{0.4\linewidth}
    \begin{table}[H]
    \centering
    \caption{Performance on Flickr under different random seeds. FedCola has the lowest standard deviation (STD).}
    \label{tab:seed}
    \resizebox{0.96\textwidth}{!}{
    \begin{tabular}{lcccc}
    \toprule
    \multirow{2.5}{*}{\bf Method}  & \multicolumn{3}{c}{\bf Seed} &  \multirow{2.5}{*}{\bf STD $\downarrow$}\\
    \cmidrule(lr){2-4}
    & 1 & 42 & 2024 & \\
    \midrule
    FedAvg  & \res{81.08} & \res{79.14} & \res{82.04} & \res{1.48} \\
    FedProx & \res{78.55} & \res{77.86} & \res{81.69} & \res{2.04} \\
    \midrule
    CreamFL & \res{74.83} & \res{75.94} & \res{78.34} & \res{1.79} \\
    FedIoT  & \res{85.51} & \res{80.16} & \res{81.10} & \res{2.86} \\
    \midrule
    FedCola & \resb{91.96} & \resb{90.80} & \resb{93.21} & \resb{1.21} \\
    \bottomrule
    \end{tabular}
    }
    \end{table}
\end{minipage}
\hfill
\begin{minipage}{0.58\linewidth}
\begin{table}[H]
    \centering
    \caption{Flickr performance under imbalanced total client numbers}
    \label{tab:imb}
    \resizebox{0.96\textwidth}{!}{
    \begin{tabular}{ccccccc}
    \toprule
     \multirow{2.5}{*}{\bf Setting} &\multirow{2.5}{*}{\bf Method}  & \multicolumn{2}{c}{\bf 1k Test Image}& \multicolumn{2}{c}{\bf 5k Test Image} &  \multirow{2.5}{*}{\bf $\mathbf{R@1_\text{sum}}$}\\
     \cmidrule(lr){3-4} \cmidrule(lr){5-6}
     & & i2t $R@1$ & t2i $R@1$ & i2t $R@1$ & t2i $R@1$ &\\
     \midrule
     \multirow{6}{*}{\makecell{\bf More \\ \bf Total Image \\ \bf Clients}}& FedAvg & \res{31.58} & \res{22.74} & \res{14.74} & \res{9.98} & \res{79.04} \\
& FedProx & \res{29.24} & \res{20.51} & \res{13.64} & \res{8.76} & \res{72.15} \\
\cmidrule{2-7}
& CreamFL & \res{29.58} & \res{21.34} & \res{13.84} & \res{9.22} & \res{73.98} \\
& FedIoT & \res{32.76} & \res{23.36} & \res{15.68} & \res{10.53} & \res{82.33} \\
\cmidrule{2-7}
& FedCola & \resb{37.16} & \resb{26.07} & \resb{18.64} & \resb{12.46} & \resb{94.33} \\

     \midrule
     \multirow{6}{*}{\makecell{\bf More \\ \bf Total Text\\ \bf Clients}}& FedAvg & \res{32.90} & \res{23.34} & \res{15.48} & \res{10.39} & \res{82.11} \\
& FedProx & \res{20.02} & \res{14.60} & \res{7.88} & \res{5.64} & \res{48.14} \\
\cmidrule{2-7}
& CreamFL & \res{30.38} & \res{21.86} & \res{13.82} & \res{9.56} & \res{75.62} \\
& FedIoT & \res{31.88} & \res{22.85} & \res{14.82} & \res{10.29} & \res{79.84} \\
\cmidrule{2-7}
& FedCola & \resb{36.24} & \resb{25.76} & \resb{17.62} & \resb{12.06} & \resb{91.68} \\

     \bottomrule
 \end{tabular}
 }
 \end{table}
\end{minipage}
\end{figure}

\begin{figure}[!t]
\begin{minipage}{0.48\linewidth}
\begin{table}[H]
    \centering
    \caption{Flickr Breakdown Performance}
    \label{tab:flickr}
    \resizebox{\linewidth}{!}{
    \begin{tabular}{ccccccc}
    \toprule
     \multirow{2.5}{*}{\bf Setting} &\multirow{2.5}{*}{\bf Method}  & \multicolumn{2}{c}{\bf 1k Test Image}& \multicolumn{2}{c}{\bf 5k Test Image} &  \multirow{2.5}{*}{\bf $\mathbf{R@1_\text{sum}}$}\\
     \cmidrule(lr){3-4} \cmidrule(lr){5-6}
     & & i2t $R@1$ & t2i $R@1$ & i2t $R@1$ & t2i $R@1$ &\\
     \midrule
     \multirow{6}{*}{\bf Default}& FedAvg& \res{32.84}& \res{22.90}& \res{15.32}& \res{10.02}& \res{81.08}\\
     & FedProx& \res{31.36}& \res{22.41}& \res{14.84}& \res{9.94}& \res{78.55}\\
     \cmidrule{2-7}
     & CreamFL& \res{30.2}& \res{21.34}& \res{13.82}& \res{9.46}& \res{74.83}\\
     & FedIoT& \res{34.42}& \res{23.87}& \res{16.34}& \res{10.88}& \res{85.51}\\
     \cmidrule{2-7}
     & FedCola& \resb{35.68}& \resb{26.14}& \resb{18.10}& \resb{12.04}& \resb{91.96}\\
     \midrule
     \multirow{6}{*}{\makecell{\bf More \\ \bf Heterogeneity}}& FedAvg& \res{32.5}& \res{23.34}& \res{15.40}& \res{10.46}& \res{81.70}\\
     & FedProx& \res{31.1}& \res{22.06}& \res{13.9}& \res{9.26}& \res{76.33}\\
     \cmidrule{2-7}
     & CreamFL& \res{31.48}& \res{22.59}& \res{15.74}& \res{10.19}& \res{80.00}\\
     & FedIoT& \res{33.02}& \res{23.73}& \res{15.94}& \res{10.57}& \res{83.28}\\
     \cmidrule{2-7}
     & FedCola& \resb{36.26}& \resb{26.06}& \resb{17.54}& \resb{11.96}& \resb{91.82}\\
     \midrule
     \multirow{6}{*}{\makecell{\bf Less \\ \bf Participation}}& FedAvg& \res{25.84}& \res{18.95}& \res{11.96}& \res{8.06}& \res{64.82}\\
     & FedProx& \res{25.94}& \res{19.02}& \res{10.74}& \res{7.64}& \res{63.33}\\
     \cmidrule{2-7}
     & CreamFL& \res{26.94}& \res{19.54}& \res{11.9}& \res{8.47}& \res{66.85}\\
     & FedIoT& \res{25.18}& \res{18.13}& \res{11.02}& \res{7.62}& \res{61.94}\\
     \cmidrule{2-7}
     & FedCola& \resb{34.94}& \resb{25.48}& \resb{16.60}& \resb{11.84}& \resb{88.85}\\
     \midrule
     \multirow{6}{*}{\makecell{\bf More \\ \bf Image}}& FedAvg& \res{31.22}& \res{22.68}& \res{14.42}& \res{9.96}& \res{78.28}\\
     & FedProx& \res{31.46}& \res{22.84}& \res{14.90}& \res{10.05}& \res{79.25}\\
     \cmidrule{2-7}
     & CreamFL& \res{32.02}& \res{23.18}& \res{14.74}& \res{10.36}& \res{80.31}\\
     & FedIoT& \res{33.22}& \res{23.40}& \res{15.78}& \res{10.34}& \res{82.74}\\
     \cmidrule{2-7}
     & FedCola& \resb{35.42}& \resb{25.8}& \resb{17.76}& \resb{12.26}& \resb{91.24}\\
     \midrule
     \multirow{6}{*}{\makecell{\bf More \\ \bf Text}}& FedAvg& \res{31.94} & \res{22.55} & \res{15.20} & \res{10.00} & \res{79.69} \\
     & FedProx& \res{31.20} & \res{22.25} & \res{14.44} & \res{9.70} & \res{77.59} \\
     \cmidrule{2-7}
     & CreamFL& \res{31.96} & \res{23.20} & \res{15.12} & \res{10.47} & \res{80.75} \\
     & FedIoT& \res{31.46} & \res{22.22} & \res{14.56} & \res{9.77} & \res{78.02} \\
     \cmidrule{2-7}
     & FedCola& \resb{35.48} & \resb{25.50} & \resb{17.40} & \resb{11.72} & \resb{90.10} \\
     \midrule
     \multirow{6}{*}{\makecell{\bf Fewer \\ \bf Image-Text}}& FedAvg& \res{24.92} & \res{18.01} & \res{10.70} & \res{7.49} & \res{61.12} \\
     & FedProx& \res{24.28} & \res{17.50} & \res{10.22} & \res{7.19} & \res{59.19} \\
     \cmidrule{2-7}
     & CreamFL& \res{23.20} & \res{17.12} & \res{9.64} & \res{7.15} & \res{57.12} \\
     & FedIoT& \res{24.68} & \res{17.76} & \res{10.42} & \res{7.47} & \res{60.34} \\
     \cmidrule{2-7}
     & FedCola& \resb{34.06} & \resb{24.28} & \resb{16.18} & \resb{11.16} & \resb{85.68} \\
     \midrule
     
    \end{tabular}
    }
\end{table}
\end{minipage}
\hfill
\begin{minipage}{0.48\linewidth}
\begin{table}[H]
    \centering
    \caption{COCO Breakdown Performance}
    \label{tab:coco}
    \resizebox{\linewidth}{!}{
    \begin{tabular}{ccccccc}
    \toprule
     \multirow{2.5}{*}{\bf Setting} &\multirow{2.5}{*}{\bf Method}  & \multicolumn{2}{c}{\bf 1k Test Image}& \multicolumn{2}{c}{\bf 5k Test Image} &  \multirow{2.5}{*}{\bf $\mathbf{R@1_\text{sum}}$}\\
     \cmidrule(lr){3-4} \cmidrule(lr){5-6}
     & & i2t $R@1$ & t2i $R@1$ & i2t $R@1$ & t2i $R@1$ &\\
     \midrule
     \multirow{6}{*}{\bf Default}& FedAvg & \res{36.98} & \res{29.28} & \res{16.76} & \res{12.40} & \res{95.42} \\
        & FedProx & \res{37.56} & \res{28.46} & \res{16.68} & \res{12.46} & \res{95.16} \\
        \cmidrule{2-7}
        & CreamFL & \res{37.60} & \res{28.64} & \res{16.68} & \res{12.34} & \res{95.26} \\
        & FedIoT & \res{38.62} & \res{29.97} & \res{17.16} & \res{12.65} & \res{98.40} \\
        \cmidrule{2-7}
        & FedCola & \resb{41.02} & \resb{31.62} & \resb{18.74} & \resb{13.72} & \resb{105.10} \\

     \midrule
     \multirow{6}{*}{\makecell{\bf More \\ \bf Heterogeneity}}& FedAvg & \res{37.46} & \res{29.11} & \res{16.40} & \res{12.35} & \res{95.32} \\
    & FedProx & \res{37.66} & \res{28.86} & \res{16.90} & \res{12.20} & \res{95.62} \\
    \cmidrule{2-7}
    & CreamFL & \res{35.76} & \res{28.11} & \res{15.74} & \res{11.80} & \res{91.41} \\
    & FedIoT & \res{37.66} & \res{29.47} & \res{16.52} & \res{12.24} & \res{95.89} \\
    \cmidrule{2-7}
    & FedCola & \resb{39.62} & \resb{30.37} & \resb{17.72} & \resb{13.12} & \resb{100.83} \\
    
     \midrule
     \multirow{6}{*}{\makecell{\bf Less \\ \bf Participation}}& FedAvg & \res{32.68} & \res{26.12} & \res{14.22} & \res{10.90} & \res{83.91} \\
& FedProx & \res{31.20} & \res{25.13} & \res{13.28} & \res{10.27} & \res{79.88} \\
\cmidrule{2-7}
& CreamFL & \res{31.58} & \res{25.00} & \res{12.60} & \res{9.79} & \res{78.97} \\
& FedIoT & \res{31.62} & \res{25.35} & \res{13.44} & \res{10.24} & \res{80.65} \\
\cmidrule{2-7}
& FedCola & \resb{40.12} & \resb{30.47} & \resb{18.28} & \resb{13.43} & \resb{102.30} \\

     \midrule
     \multirow{6}{*}{\makecell{\bf More \\ \bf Image}}& FedAvg & \res{38.26} & \res{29.33} & \res{17.22} & \res{12.47} & \res{97.28} \\
& FedProx & \res{37.46} & \res{28.67} & \res{16.80} & \res{12.46} & \res{95.39} \\
\cmidrule{2-7}
& CreamFL & \res{36.80} & \res{28.66} & \res{16.02} & \res{12.17} & \res{93.65} \\
& FedIoT & \res{36.86} & \res{29.06} & \res{16.78} & \res{12.34} & \res{95.04} \\
\cmidrule{2-7}
& FedCola & \resb{40.58} & \resb{31.05} & \resb{19.26} & \resb{13.33} & \resb{104.22} \\

     \midrule
     \multirow{6}{*}{\makecell{\bf More \\ \bf Text}}& FedAvg & \res{38.00} & \res{28.95} & \res{17.26} & \res{12.48} & \res{96.69} \\
& FedProx & \res{36.96} & \res{28.70} & \res{16.92} & \res{12.38} & \res{94.96} \\
\cmidrule{2-7}
& CreamFL & \res{36.68} & \res{28.46} & \res{15.62} & \res{12.06} & \res{92.81} \\
& FedIoT & \res{37.74} & \res{29.47} & \res{17.22} & \res{12.61} & \res{97.04} \\
\cmidrule{2-7}
& FedCola & \resb{39.82} & \resb{30.32} & \resb{17.96} & \resb{12.86} & \resb{100.96} \\
     \midrule
     \multirow{6}{*}{\makecell{\bf Fewer \\ \bf Image-Text}} & FedAvg & \res{30.30} & \res{23.78} & \res{12.02} & \res{8.99} & \res{75.10} \\
& FedProx & \res{29.22} & \res{23.32} & \res{11.32} & \res{9.22} & \res{73.08} \\
\cmidrule{2-7}
& CreamFL & \res{29.60} & \res{23.78} & \res{12.18} & \res{9.12} & \res{74.69} \\
& FedIoT & \res{30.80} & \res{23.56} & \res{12.42} & \res{9.36} & \res{76.14} \\
\cmidrule{2-7}
& FedCola & \resb{37.78} & \resb{28.08} & \resb{16.80} & \resb{11.74} & \resb{94.40} \\

     \midrule
     
    \end{tabular}
}
\end{table}
\end{minipage}
\end{figure}

\begin{figure}[!t]
\begin{subfigure}[t]{0.32\textwidth}
    \centering
    \includegraphics[width=\linewidth]{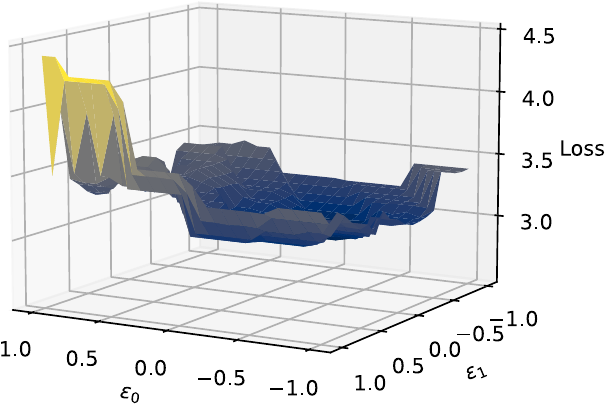}
    \caption{FedAvg}
    \label{fig:fedavg}
\end{subfigure}
\hfill
\begin{subfigure}[t]{0.32\textwidth}
    \centering
    \includegraphics[width=\linewidth]{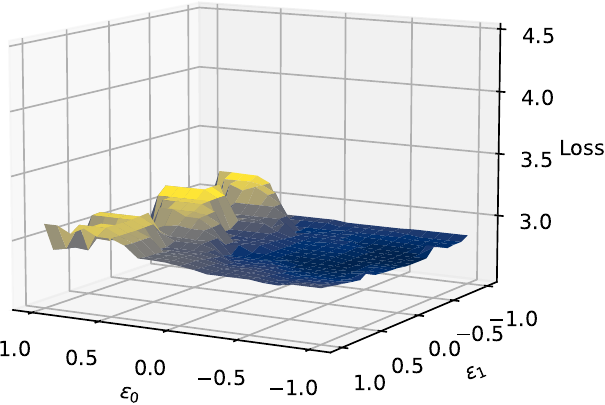}
    \caption{FedCola}
    \label{fig:fedcola}
\end{subfigure}
\begin{subfigure}[t]{0.33\textwidth}
    \centering
    \includegraphics[width=\linewidth]{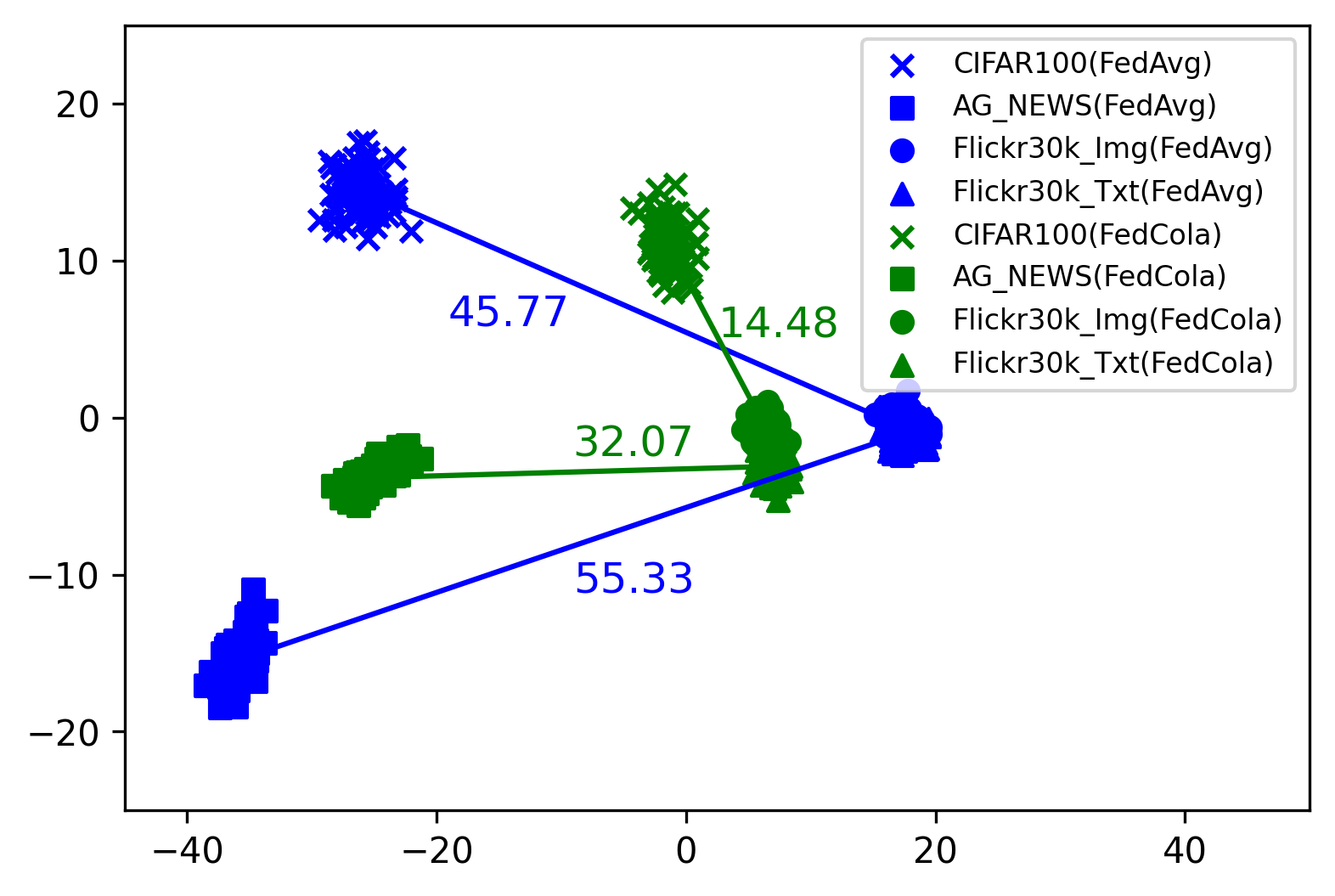}
    \caption{Extracted features}
    \label{fig:features}
\end{subfigure}
\caption{ Visualization of the parametric loss landscape with Hessian eigenvectors $\epsilon_0$ and $\epsilon_1$ and the extracted features for each resulting global multi-modal model.}
\end{figure}

\subsection{Visualization}

The smoothness of the parametric loss space has been utilized as a significant indicator of the model generalizabilty~\cite{fedalign,chen2020stabilizing,zela2019understanding}. To illustrate that FedCola learns a more generalized global model, we visualize the loss space on $256$ training samples of FedAvg (Fig.~\ref{fig:fedavg}) and FedCola (Fig.~\ref{fig:fedcola}) when the weights of the model are perturbed along the direction of the top Hessian eigenvectors. The loss landscape of FedCola is significantly smoother than FedAvg, indicating that with the help of the proposed framework, a more generalized global model can be obtained. Additionally, we further conduct visualizations with Linear Discriminant Analysis (LDA) at the feature level, as shown in Fig.~\ref{fig:features}. By computing the distance between the feature centers, we find the gaps between uni-modal and multi-modal datasets are reduced under FedCola. 

\section{Potential Negative Societal Impact and Limitation}\label{sec:lim}
\subsubsection{Potential Negative Societal Impact.} The effectiveness of FedCola, like any machine learning model, is contingent on the data it's trained on. Given that data distribution in FL settings can be highly non-uniform and biased towards certain demographics or modalities, there's a risk of amplifying existing biases or creating new ones. This can lead to unfair models that perform inequitably across different groups or modalities.
% \nopagebreak

\subsubsection{Limitations.} FedCola currently does not address system heterogeneity, representing a limitation in the present framework. We propose to explore this aspect in future research.

\end{document}